\documentclass[journal]{IEEEtran}
\usepackage[pdftex]{graphicx}
\usepackage{acronym}
\usepackage{multirow}
\usepackage{amsmath, amsthm, amssymb, amsfonts, mathtools}
\usepackage{placeins}
\usepackage[ruled,vlined]{algorithm2e}
\usepackage{algorithmic}
\usepackage{url}
\usepackage{hhline}
\usepackage[stable]{footmisc}
\renewcommand{\baselinestretch}{1.0}\normalsize

\usepackage{tikz}
\usepackage{pgfplots}
\usepackage{times}
\usetikzlibrary{calc}
\usetikzlibrary{backgrounds}
\providecommand{\mat}[1]{\boldsymbol{#1}}
\providecommand{\vct}[1]{\boldsymbol{#1}}
\usepackage{soul}

\usepackage{xcolor}
\renewcommand{\refeq}[1]{{Eq.~(\ref{#1})}}

\newcommand{\reffig}[1]{{Fig.~\ref{#1}}}

\begin{document}
%
\title{Online Few-shot Gesture Learning on a Neuromorphic Processor}

\providecommand{\keywords}[1]
{
  \textbf{\textit{Keywords---}} #1
}

\acrodef{AC}[AC]{Arrenhius \& Current}
\acrodef{AD}[AD]{Automatic Differentiation}
\acrodef{ANN}[ANN]{Artificial Neural Network}
\acrodef{AER}[AER]{Address Event Representation}
\acrodef{AEX}[AEX]{AER EXtension board}
\acrodef{AMDA}[AMDA]{``AER Motherboard with D/A converters''}
\acrodef{API}[API]{Application Programming Interface}
\acrodef{BP}[BP]{Back-Propagation}
\acrodef{BPTT}[BPTT]{Back-Propagation-Through-Time}
\acrodef{BM}[BM]{Boltzmann Machine}
\acrodef{CAVIAR}[CAVIAR]{Convolution AER Vision Architecture for Real-Time}
\acrodef{CCN}[CCN]{Cooperative and Competitive Network}
\acrodef{CD}[CD]{Contrastive Divergence}
\acrodef{CMOS}[CMOS]{Complementary Metal--Oxide--Semiconductor}
\acrodef{COTS}[COTS]{Commercial Off-The-Shelf}
\acrodef{CPU}[CPU]{Central Processing Unit}
\acrodef{CV}[CV]{Coefficient of Variation}
\acrodef{CV}[CV]{Coefficient of Variation}
\acrodef{DAC}[DAC]{Digital--to--Analog}
\acrodef{DBN}[DBN]{Deep Belief Network}
\acrodef{DCLL}[DECOLLE]{Deep Continuous Local Learning}
\acrodef{DFA}[DFA]{Deterministic Finite Automaton}
\acrodef{DFA}[DFA]{Deterministic Finite Automaton}
\acrodef{divmod3}[DIVMOD3]{divisibility of a number by 3}
\acrodef{DPE}[DPE]{Dynamic Parameter Estimation}
\acrodef{DPI}[DPI]{Differential-Pair Integrator}
\acrodef{DSP}[DSP]{Digital Signal Processor}
\acrodef{DVS}[DVS]{Dynamic Vision Sensor}
\acrodef{EDVAC}[EDVAC]{Electronic Discrete Variable Automatic Computer}
\acrodef{EIF}[EI\&F]{Exponential Integrate \& Fire}
\acrodef{EIN}[EIN]{Excitatory--Inhibitory Network}
\acrodef{EOS}[EOS]{Error-Triggered Online Surrogate gradient}
\acrodef{EPSC}[EPSC]{Excitatory Post-Synaptic Current}
\acrodef{EPSP}[EPSP]{Excitatory Post--Synaptic Potential}
\acrodef{eRBP}[eRBP]{Event-Driven Random Back-Propagation}
\acrodef{FPGA}[FPGA]{Field Programmable Gate Array}
\acrodef{FSM}[FSM]{Finite State Machine}
\acrodef{GPU}[GPU]{Graphical Processing Unit}
\acrodef{HAL}[HAL]{Hardware Abstraction Layer}
\acrodef{HH}[H\&H]{Hodgkin \& Huxley}
\acrodef{HMM}[HMM]{Hidden Markov Model}
\acrodef{HW}[HW]{Hardware}
\acrodef{hWTA}[hWTA]{Hard Winner--Take--All}
\acrodef{IF2DWTA}[IF2DWTA]{Integrate \& Fire 2--Dimensional WTA}
\acrodef{IF}[I\&F]{Integrate \& Fire}
\acrodef{IFSLWTA}[IFSLWTA]{Integrate \& Fire Stop Learning WTA}
\acrodef{INCF}[INCF]{International Neuroinformatics Coordinating Facility}
\acrodef{INI}[INI]{Institute of Neuroinformatics}
\acrodef{IO}[IO]{Input-Output}
\acrodef{IPSC}[IPSC]{Inhibitory Post-Synaptic Current}
\acrodef{ISI}[ISI]{Inter--Spike Interval}
\acrodef{JFLAP}[JFLAP]{Java - Formal Languages and Automata Package}
\acrodef{LIF}[LI\&F]{Leaky Integrate \& Fire}
\acrodef{LSM}[LSM]{Liquid State Machine}
\acrodef{LSTM}[LSTM]{Long Short Term Memory}
\acrodef{LTD}[LTD]{Long-Term Depression}
\acrodef{LTI}[LTI]{Linear Time-Invariant}
\acrodef{LTP}[LTP]{Long-Term Potentiation}
\acrodef{LTU}[LTU]{Linear Threshold Unit}
\acrodef{MCMC}{Markov Chain Monte Carlo}
\acrodef{NHML}[NHML]{Neuromorphic Hardware Mark-up Language}
\acrodef{NMDA}[NMDA]{NMDA}
\acrodef{NME}[NE]{Neuromorphic Engineering}
\acrodef{PCB}[PCB]{Printed Circuit Board}
\acrodef{PRC}[PRC]{Phase Response Curve}
\acrodef{PSC}[PSC]{Post-Synaptic Current}
\acrodef{PSP}[PSP]{Post--Synaptic Potential}
\acrodef{RI}[KL]{Kullback-Leibler}
\acrodef{RRAM}[RRAM]{Resistive Random-Access Memory}
\acrodef{RTRL}[RTRL]{Real-Time Recurrent Learning}
\acrodef{RBM}[RBM]{Restricted Boltzmann Machine}
\acrodef{ROC}[ROC]{Receiver Operator Characteristic}
\acrodef{SAC}[SAC]{Selective Attention Chip}
\acrodef{SCD}[SCD]{Spike-Based Contrastive Divergence}
\acrodef{SCX}[SCX]{Silicon CorteX}
\acrodef{SG}[SG]{Surrogate Gradient}
\acrodef{SRM}[SRM$_0$]{Spike Response Model}
\acrodef{SNN}[SNN]{Spiking Neural Network}
\acrodef{RNN}[RNN]{Recurrent Neural Network}
\acrodef{SOEL}[SOEL]{Surrogate-gradient  Online Error-Triggered Learning}
\acrodef{STDP}[STDP]{Spike Time Dependent Plasticity}
\acrodef{SW}[SW]{Software}
\acrodef{sWTA}[SWTA]{Soft Winner--Take--All}
\acrodef{VHDL}[VHDL]{VHSIC Hardware Description Language}
\acrodef{VLSI}[VLSI]{Very  Large  Scale  Integration}
\acrodef{WTA}[WTA]{Winner--Take--All}
\acrodef{XML}[XML]{eXtensible Mark-up Language}

\author{\IEEEauthorblockN{Kenneth Stewart}
\IEEEauthorblockA{Department of Computer Science \hspace{-5.5mm}\\
UC, Irvine\\
Irvine, CA 92697-2625 USA\\
kennetms@uci.edu}\\
\and
\IEEEauthorblockN{Garrick Orchard}
\IEEEauthorblockA{Intel Labs\\Intel Corporation\\
Santa Clara, California\\
garrick.orchard@intel.com}\\
\and
\IEEEauthorblockN{Sumit Bam Shrestha}
\IEEEauthorblockA{Institute for Infocomm Research \\
A*STAR\\
Singapore 138632\\
Sumit\_Bam@i2r.a-star.edu.sg}\\
\and
\IEEEauthorblockN{Emre Neftci}
\IEEEauthorblockA{ Department of Cognitive Sciences,\\Department of Computer Science\\
UC, Irvine\\
Irvine, CA 92697-2625 USA\\
eneftci@uci.edu}
}

\maketitle


\newif\ifblacktext
\blacktextfalse

\ifblacktext
\newcommand{\go}[1]{\bf GO: #1} 
\newcommand{\eon}[1]{\bf EON: #1} 
\newcommand{\ks}[1]{\bf KS: #1} 
\newcommand{\sbs}[1]{\bf SBS: #1} 
\else
\newcommand{\go}[1]{\textcolor{orange}{GO: #1}} 
\newcommand{\eon}[1]{\textcolor{blue}{EN: #1}} 
\newcommand{\ks}[1]{\textcolor{teal}{KS: #1}} 
\newcommand{\sbs}[1]{\textcolor{green}{SBS: #1}} 
\fi

\begin{abstract}
    We present the Surrogate-gradient Online Error-triggered Learning (SOEL) system for online few-shot learning on neuromorphic processors. The SOEL learning system uses a combination of transfer learning and principles of computational neuroscience and deep learning. We show that partially trained deep \acp{SNN} implemented on neuromorphic hardware can rapidly adapt online to new classes of data within a domain. SOEL updates trigger when an error occurs, enabling faster learning with fewer updates. Using gesture recognition as a case study, we show SOEL can be used for online few-shot learning of new classes of pre-recorded gesture data and rapid online learning of new gestures from data streamed live from a Dynamic Active-pixel Vision Sensor to an Intel Loihi neuromorphic research processor. 
\end{abstract}

\hspace{8pt}


\section{Introduction}
The current generation of \acp{ANN} achieve state of the art performance in applications ranging from image classification and object recognition, to object tracking, signal processing, natural language processing, self driving cars, health care diagnostics, and many more \cite{Goodfellow_etal16_deeplear}. 
\acp{ANN} mainly rely on the backpropagation of errors as the key to their learning prowess, but they require large amounts of data and memory for training. 
Training these networks relies on GPUs and thousands of iterations over the data sampled in an i.i.d. fashion. 
To achieve the high throughput necessary to quickly train the networks, GPUs rely on high volumes of data movement and tensor-based computations, both of which consume large amounts of energy \cite{nagasaka2010pow,kestor2013datamv}, making GPU less than ideal for implementation at scale in mobile systems.

Neuromorphic computing platforms offer an energy-efficient alternative to perform training and inference in neural networks while being suitable for power-constrained applications in mobile systems \cite{hwu2017complete}. 
Neuromorphic systems mimic the brain's event-driven dynamics, distributed architecture and massive parallelism to overcome the limitations of conventional von Neumann computing architectures \cite{Indiveri_etal11_neursili}. 
Neuromorphic hardware equipped with synaptic plasticity capability can perform training and inference online, using local information \cite{Chicca_etal13_neurelec,Davies_etal18_loihneur}, making them particularly interesting for problems requiring fast adaptation to new data.
Despite the many technical advances in neuromorphic learning hardware, the practical role of learning and synaptic plasticity in neuromorphic hardware has remained elusive. This is because gradient-based learning is notoriously slow, requiring many iterations to achieve an acceptable generalization. Furthermore, synaptic plasticity is inherently online and local, but learning online using streaming data breaks the i.i.d. assumptions required for convergence of the neural network.

In this paper, we take a realistic and practical approach to learning in neuromorphic hardware by combining the best of conventional and neuromorphic hardware: we pre-train networks on GPUs on a class of tasks and adapt the key layers on the neuromorphic hardware. The result is a system that moves the non-local and energy-intensive phases of learning to a cloud or mainframe, and deploys on the neuromorphic hardware the data sensitive portions of the learning.
This approach is realized by using recent theories that combine machine learning theory with \acp{SNN} and few-shot/transfer learning.
We demonstrate our approach on fast online learning of streaming, real-world visual patterns on a neuromorphic processor. 
The pre-training is carried out using a functional model of an Intel Loihi neuromorphic processor. 
During deployment, the model is then fine-tuned on the processor using local synaptic plasticity rules. 
The key contribution of this work is few-shot \ac{SOEL}, a plasticity rule compatible with neuromorphic hardware derived from gradient-descent on \acp{SNN} and its efficient implementation using signals local to the neuromorphic cores.

\subsection{\acf{SOEL}}
While gradient \ac{BP} is the workhorse for training nearly all deep neural network architectures, it is generally incompatible with non Von Neumann computers, including brains and neuromorphic hardware.
By identifying \acp{SNN} as a type of \ac{RNN}, recent studies showed two reasons for this incompatibility \cite{Neftci_etal19_surrgrad}. 
Firstly, the spiking neuron has a non-differentiable activation function, which prevents the gradients from flowing across the network. 
Secondly, the computation of local errors requires the evaluation of a global loss function, which is a spatially and temporally non-local computation.

These problems are addressed using \ac{SG} learning~\cite{Neftci_etal19_surrgrad}.
\ac{SG} methods define a differentiable surrogate network to calculate weight updates in a local fashion, and formulate the updates as three-factor synaptic plasticity rules. 
The \ac{SG} method reveals from first principles the mathematical nature of the three factors, and a learning dynamic that is temporally continuous and compatible with synaptic plasticity.
The three factor rules include a pre-synaptic factor, a post-synaptic factor and an external error signal. 
In comparison, \ac{STDP}, a common synaptic plasticity model in neuroscience, contains only two factors and lacks the external signal \cite{Bi_Poo98_synamodi}. 
The third factor drastically improves learning by projecting task-specific errors to the neurons \cite{Baldi_etal17_learmach}.
In short, the \ac{SG} bridges the worlds of \acp{ANN} and \acp{SNN} without simplifying assumptions on the latter.

\ac{SG} methods pave  the  road  towards  neuromorphic  learning  machines  with  performances  similar  to  deep-learning \cite{Kaiser_etal20_synaplas},  while  being  able  to  learn  online,  with  input  streams, spike timing  and potentially using a fraction of the energy compared to conventional computers.

While temporal continuity is a plausible property in the brain, updating a large number of weights continuously is both energetically expensive and prone to dynamical instabilities. 
A recent development of \ac{SG} learning in spiking neurons suggested that updating at every timestep is not necessary if weight updates are triggered by task errors \cite{Payvand_etal20_errothre}. 
In such \emph{error-triggered} learning, weight updates are made only when an error threshold is crossed. Consequently, the number of updates can be drastically reduced with a small penalty in final accuracy. 

However, even the fewer error-triggered learning updates is incompatible with online learning as it can lead to catastrophic forgetting.
Catastrophic forgetting occurs when the data generating process for training the neural network is non i.i.d.
This problem can be generally solved by increasing the complexity of the neuron and synapses \cite{Zenke_etal17_imprmult,Kirkpatrick_etal17_overcata}, experience replays \cite{Mnih_etal15_humacont}, or meta-learning and the related few-shot learning \cite{Andrychowicz_etal16_learto}.

Here, we focus on the latter approach for the following reasons: Firstly, the ability to solve difficult recognition tasks using few samples is a key capability of the brain \cite{Lake_etal15_humaconc}.
Few-shot learning is a subset of deep learning concerned with such fast adaptation in situations where prior knowledge of the task domain is available. 
Bringing such capability to neuromorphic hardware is a top priority for local learning and adaptation on mobile systems, such as the learning of human gestures for device control or adapting to a user's voice.

In our approach, few-shot learning consists of first pre-training a model on the class of problems of interest, and then making (presumably) few error-triggered updates to learn new but related tasks.

To achieve the error-triggered learning in neuromorphic hardware, we implement the \ac{SOEL} algorithm, an extension of \ac{SG} and error-triggered learning for rapid, few-shot learning.
We demonstrate \ac{SOEL} on the Intel Loihi Neuromorphic research chip, and capitalized on its specialized local plasticity processors to carry out the updates.
\ac{SOEL} is an extension of the Surrogate Gradient learning algorithm designed that fix issues of a previous implementation~\cite{Stewart_etal20_on-cfew-}.

\section{Related Work}
Previous work has shown that the first layers of neural networks learn general features and learn increasingly task specific features the deeper within the network the layer is \cite{Yosinski_etal14_howtran}.
The general features learned by the first layers of a network can be transferred to other networks for task-specific training of later layers, referred to as transfer learning.

The first layers of a network are trained on one dataset to learn general features that can be transferred to a second network trained on a target task. 
Using the transferred features yields better generalization of the target task than without transferring the general lower layer features \cite{Yosinski_etal14_howtran}. 
Transfer learning is useful for few-shot learning, \emph{i.e.} when the target task only has a small amount of data available for training, but a similar task with a larger dataset is available.
The goal of few-shot learning is to train a model to generalize from as few examples as possible \cite{Vinyals_etal16_matcnetw}.
Transfer learning can be used to assist the training of few-shot learning models allowing for greater generalization on a target domain from few examples for both \acp{ANN} and deep \acp{SNN} \cite{Qiao2018fewshot,scott2018kshottransfer,Stewart_etal20_on-cfew-}. 

As discussed earlier, training deep \acp{SNN} is challenging due to a spatiotemporal credit assignment problem and non-differentiabilities.
Previous work overcame the learning problem in multiple layers of \acp{SNN} with methods such as feedback alignment \cite{Neftci_etal17_evenranda,Lillicrap_etal16_randsyna}, backpropagation-through-time (BPTT) \cite{Shrestha_Orchard18_slayspik,yin2020effective,Huh_Sejnowski17_graddesc}, and spike-based backpropagation \cite{Thiele_etal19_spikann-,Lee2020spikebp}.
The successful gradient-based training of deep \acp{SNN} usually approximate the spiking function's derivative using a surrogate activation function \cite{Neftci_etal19_surrgrad}.
By being able to train deep \acp{SNN}, the ingredients of deep learning that make \acp{ANN} successful such as dropout, batch normalization, convolutions, pooling etc. can be applied to \acp{SNN}, in addition to \acp{SNN} being compatible with neuromorphic hardware.
BPTT is shown to achieve state-of-the-art accuracy on certain target domains such as NMNIST and gesture recognition. 
However, BPTT is inherently offline as it propagates error through the unrolled network and therefore is not suited for applications where online adaptation is desired. 
Rather than backpropagating through the network, \ac{SOEL} computes local errors between pre- and post-synaptic neurons by propagating gradients forward in time \cite{Neftci_etal19_surrgrad}. 

While previous work has shown increasing success in training \acp{SNN} using spike-based gradient descent on a variety of tasks, they are trained and tested offline just like \acp{ANN} and do not demonstrate online on-chip learning on neuromorphic hardware. \cite{Imam2020smell} demonstrated rapid online on-chip learning using the Intel Loihi neuromorphic research chip but did not use spike-based gradient descent. 
To our knowledge this work is the first to demonstrate online, on-chip gradient-based learning on a neuromorphic processor.
Using gestures as a case study we show the success of rapid online learning using \emph{SOEL} which can be used for applications that require online adaptation. 


\section{Background}
\subsection{Dynamic Vision Sensor}
The datasets used in this work are obtained using neuromorphic sensors, namely the DVS and DAVIS cameras.
Each pixel of Dynamic Vision Sensors (DVSs) quantize local relative intensity changes to generate spike events \cite{Lichtsteiner_etal08_128x120}. In our experiments we use data from a DVS 128, and a  DAVIS 240C \cite{Brandli_etal14_240180}.
The IBM DvsGesture dataset used here for pre-training consists of recordings of 29 different individuals performing 10 different actions such as clapping and an unspecified gesture for a total of 11 classes. The actions are recorded  using a DVS camera, an event-based neuromorphic sensor, under three different lighting conditions. The task is to classify an action sequence video. Samples from the first 23 subjects were used for training and the last 6 subjects were used for testing. The training set contains 1078 samples and the test set contains 264 samples. Each sample consists of the first 1.45 seconds of the gesture performed.

\subsection{Neural Network Model}
The neural network model follows leaky, \ac{IF} dynamics.
The dynamics of the membrane potential $U_i$ of a neuron $i$ is governed by the following differential equations:
\begin{equation}\label{eq:clif}
  \begin{split}
    U_i(t) =& V_i(t) - U_{th} R_i(t) + b_i,\\
    \tau_{mem}\frac{\mathrm{d}}{\mathrm{d}t} V_i(t) =& - V_i(t) + I_i(t),\\
    \tau_{ref} \frac{\mathrm{d}}{\mathrm{d}t} R_i(t) =& -R_i(t)  +  S_i(t),\\
  \end{split}
\end{equation}
\noindent with $S_i(t)=\sum_f \delta (t-t_i^f)$  representing the spike train of neuron $i$ spiking at times $t^f_i$, where $\delta$ is the Dirac delta.
A spike is emitted when the membrane potential reaches a threshold $U_{th}$.

The constant $b_i$ represents the intrinsic excitability of the neuron.
The reset mechanism is captured with the dynamics of $R_i$.
The factors $\tau_{mem}$ and $\tau_{ref}$ are time constants of the membrane and reset dynamics, respectively.
$I_{i}$ denotes the total synaptic current of neuron $i$, expressed as:
\begin{equation}\label{eq:Iv}
    \begin{split}
      \tau_{syn} \frac{\mathrm{d}}{\mathrm{d} t} I_{i}(t)  = & -I_{i}(t) + \sum_{j\in \text{pre}} W_{ij} S_j(t),
    \end{split}
\end{equation}
\noindent where $W_{ij}$ is the synaptic weights between pre-synaptic neuron $j$ and post-synaptic neuron $i$.
Because $V_i$ and $I_i$ are linear with respect to the weights $W_{ij}$, the dynamics of $V_i$ can be rewritten as:
\begin{equation}\label{eq:pq}
    \begin{split}
      V_i(t) =& \sum_{j\in \text{pre}} W_{ij} P_{j}(t), \\
      \tau_{mem} \frac{\mathrm{d}}{\mathrm{d} t} P_{j}(t)  = & -P_{j}(t) +  Q_{j}(t),\\
      \tau_{syn} \frac{\mathrm{d}}{\mathrm{d} t} Q_{j}(t)  = & -Q_{j}(t) +  S_{j}(t).
    \end{split}
\end{equation}
The states $P$ and $Q$ describe the traces of the membrane and the current-based synapse, respectively.
For each incoming spike, each trace undergoes a jump of height $1$ and otherwise decays exponentially with a time constant $\tau_{\mathrm{mem}}$ (for $P$) and $\tau_{\mathrm{syn}}$ (for $Q$). 
Weighting the trace $P_{j}$ with the synaptic weight $W_{ij}$ results in the \acp{PSP} of neuron $i$ caused by input neuron $j$.

\subsubsection*{Discrete Spike Response Model of the Neuron and Synapse Dynamics}

In a digital system, the continuous dynamics above are simulated in discrete time, with time step $\Delta t$.
The dynamical equations in \refeq{eq:clif} and \refeq{eq:pq} are expressed in discrete time as:
\begin{equation}\label{eq:lif_equations}
  \begin{split}
    U_i[t] &= \sum_j W_{ij} P_j[t] - U_{th} R_i[t]  + b_i, \\
    S_i[t] &= \Theta( U_i[t]), \\
    P_j[t+\Delta t] &= \alpha P_{j}[t] + (1-\alpha) Q_{j}[t], \\
    Q_j[t+\Delta t] &= \beta  Q_{j}[t] + (1-\beta ) S_{j}[t]
  \end{split}
\end{equation}
\noindent where the constants $\alpha=\exp(-\frac{\Delta t}{\tau_{\mathrm{mem}}})$ and $\beta=\exp(-\frac{\Delta t}{\tau_{\mathrm{syn}}})$ reflect the decay dynamics of the membrane potential $U$ and the synaptic state $I$ during a $\Delta t$ timestep.
$\Theta(U_i[t])$ is the unit step function, where $\Theta(U_i) = 0$ if $U_i < U_{th}$, otherwise $1$.
Note that \refeq{eq:lif_equations} is equivalent to a discrete-time version of the \ac{SRM} with linear filters \cite{Gerstner_Kistler02_spikneur}.

\subsection{Gradient-based Training of \acp{SNN}}
\label{sec:surrogateGradient}
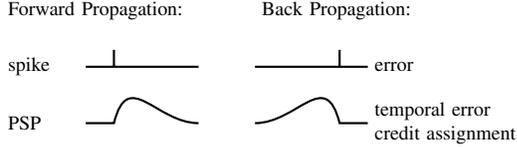
\begin{figure}[!ht]
    \centering
    \begin{tikzpicture}[scale=0.75, every node/.style={transform shape}]
    \begin{scope}
    	\node[anchor = west] at (-1.5, 1) {Forward Propagation:};
    	\begin{scope}
    		\node[anchor = west] at (-1.5, 0) {spike};
    		\draw[thick] (0, 0) -- ++ (2, 0) (0.5, 0) -- ++(0, 0.3); 
    	\end{scope}
    	
    	\begin{scope}[shift = {(0, -1)}]
    		\node[anchor = west] at (-1.5, 0) {PSP};
    		\draw[thick] (0, 0) -- (0.5, 0) .. controls (0.75, 1) and (1.25, 0) .. (2, 0);
    	\end{scope}
    	
    \end{scope}
    
    \begin{scope}[shift = {(3, 0)}]
    	\node[anchor = west] at (0, 1) {Back Propagation:};
    	\begin{scope}
    		\node[anchor = west] at (2, 0) {error};
    		\draw[thick] (0, 0) -- ++ (2, 0) (1.5, 0) -- ++(0, 0.3); 
    	\end{scope}
    	
    	\begin{scope}[shift = {(0, -1)}]
    		\node[anchor = west, text width = 2.75cm] at (2, 0) {temporal error credit assignment};
    		\draw[thick] (2, 0) -- (1.5, 0) .. controls (1.25, 1) and (0.75, 0) .. (0, 0);
    	\end{scope}
    	
    \end{scope}
    \end{tikzpicture}
    \caption{Temporal credit assignment of an error at a point in time during SLAYER backpropagation.}
    \label{fig:temporalCredit}
\end{figure}

A number of recent methods for training \acp{SNN} using gradient descent have recently emerged. 
The mathematical principle of gradient descent lies on incrementally updating the parameters in the direction opposite to the gradient of a loss function.
As mentioned in the introduction, difficulties of training \acp{SNN} arise due to the spatiotemporal credit assignment problem and the non-differentiability of the activation function. 
The spatial credit assignment problem arises when the parameters of neurons with no direct target are trained. 
The temporal credit assignment arises organically due to the temporal dependencies (dynamics) of spiking neurons.
For example, the effect of an input spike at a particular time affects the membrane potential of a spiking neuron in the future. 
The magnitude of the effect is determined by the \textit{normalized post-synaptic response} of the synapse, \emph{i.e.} the $P$ traces in \refeq{eq:lif_equations}. 
As a consequence, during learning, the credit of the error at a given point of time must be assigned to the input synapse at some point in the past. 
One factor to the magnitude of this temporal credit assignment is proportional to the normalized post-synaptic response, reversed in time. 
An illustration of this temporal credit assignment policy is shown in \reffig{fig:temporalCredit}. 
\begin{figure}[!ht]
\centering
\begin{tikzpicture}[scale=0.75, every node/.style={transform shape}]
    \def\h{0.5}
    \def\dh{0.2}
    \def\g{1}
    \def\dg{0.2}
    \def\r{0.3}
    \definecolor{fwdColor}{rgb}{0.12, 0.47, 0.71}
    \definecolor{bwdColor}{rgb}{0.8, 0.2, 0.05}
    \definecolor{prmColor}{rgb}{0.05, 0.8, 0.2}

    \begin{scope}
      \node[fwdColor] (spikeIn) at (0, 0) {$\vct S_\text{in}[t]$};
        \node[fwdColor] (weightedSpikes) at (2, 0) {
            \parbox{3em}{\centering \small Synaptic\\Weight}
        };
        \node[fwdColor] (weightsQ) at ($(weightedSpikes) + (0, 2)$) {\small Quantize};
        \node[fwdColor] (weights)  at ($(weightsQ) + (0, 1)$) {$\mat W$};
        \node[fwdColor!80] (spikeResponse) at ($(weightedSpikes) + (3, 0)$) {
            \parbox{3.25em}{\centering \small Spike\\Response}
        };
        \node[fwdColor!80] (spike) at ($(spikeResponse) + (3, 0)$) {\small Spike};
        \node[fwdColor] (params) at ($0.5*(spikeResponse.east) + 0.5*(spike.west) + (0, 0.5*\h)$) {\small Loihi Neuron};
        \node[fwdColor] (params) at ($0.5*(spikeResponse) + 0.5*(spike) + (0, 1.2)$) {\small Neuron Parameters};
        \node[fwdColor] (spikeOut) at ($(spike) + (2, 0)$) {$\vct S[t]$};

        \draw[fwdColor, ->, thick] (spikeIn) -- (weightedSpikes);
        \draw[fwdColor, ->, thick] (weightedSpikes) -- (spikeResponse) node[above, pos=0.45] {$\mat W\!\!_Q\vct S[t]$};
        \draw[fwdColor!80, ->, thick] (spikeResponse) -- (spike) node[near start, below] {$\vct U[t]$};
        \draw[fwdColor, ->, thick] (spike) -- (spikeOut);
        \draw[fwdColor, ->, thick] (weights) ++ (0, -\r) -- (weightsQ);
        \draw[fwdColor, ->, thick] (params) -- ($0.5*(spikeResponse) + 0.5*(spike) + (0, \h)$);
        \draw[fwdColor, ->, thick] (weightsQ) -- ($(weightedSpikes) + (0, \h)$) node[left, pos=0.5] {$\mat W\!\!_Q$};

        \draw[bwdColor, ->, thick] 
            ($(spikeOut.west) - (0, \g)$) -| ($(spike.east) - (\dg, \h-\dh)$) 
            node[pos=0.05, anchor=west] {$\frac{\partial\mathcal{L}(\vct S^N)}{\partial{\vct S[t]}}$};
        \draw[bwdColor, ->, thick]
            ($(spike.west) + (\dg,-\h+\dh)$) -- ++(0, -\g+\h-\dh) --
            ($(spikeResponse.east) - (\dg, \g)$) 
            node[below, pos=0.5] {\small surrogate gradient} --
            ($(spikeResponse.east) - (\dg, \h-\dh)$);
        \draw[bwdColor, ->, thick]
            ($(spikeResponse.west) + (\dg,-\h+\dh)$) -- ++(0, -\g+\h-\dh) --
            ($(weightedSpikes.east) - (\dg, \g)$) 
            node[below, pos=0.5] {\small \begin{tabular}{c}temporal credit\\assignment\end{tabular}} --
            ($(weightedSpikes.east) - (\dg, \h)$);
        \draw[bwdColor, ->, thick]
            ($(weightedSpikes.west) + (\dg,-\h)$) -- ++(0, -\g+\h) --
            ($(spikeIn.east) - (0, \g)$)
            node[anchor=east, pos=0.9] {$\frac{\partial\mathcal{L}(\vct S^N)}{\partial{\vct S_\text{in}}[t]}$};
        \draw[bwdColor, ->, thick] ($(weightedSpikes) - (0, \h)$) -- ++ (0, -1.5) 
            node[anchor=north] (gradW) {$\frac{\partial\mathcal{L}(\vct S^N)}{\partial\mat W}$};
        \end{scope}
    \begin{scope}[on background layer]
        \fill[fwdColor, opacity=0.05] ($(spikeIn) - (0.6, \h-\dh)$) rectangle ($(spikeOut) + (1,  3.7)$);
        \fill[bwdColor, opacity=0.05] ($(spikeIn) - (0.6, \h-\dh)$) rectangle ($(spikeOut) + (1, -3)$);
        \fill[prmColor, opacity=0.20] ($(params)  - (5.25, 0.4)$) rectangle ($(params) + (1.5, 0.4)$);
        \fill[gray    , opacity=0.20] ($(weights) - (0.5, 0.4)$) rectangle ($(weights) + (0.5, 0.4)$);
        \fill[prmColor, opacity=0.15] 
            ($(params) + (1.5, 0.4)$) -- ++(0, 0.2) arc(0:90:0.1) -- ++(-2.4, 0) arc(90:180:0.1) -- ++(0,-0.1) arc(0:-90:0.1);
        \fill[gray    , opacity=0.15] 
            ($(weights) + (0.5, 0.4)$) -- ++(2.2, 0) arc(90:0:0.1) -- ++(0, -0.15) arc(0:-90:0.1) -- ++(-2.1, 0) arc(90:180:0.1);

        \node[fwdColor, anchor=south west, rotate=90] at (11, -0.44) {\small Loihi Functional Simulation};
        \node[bwdColor, anchor=south west, rotate=90] at (11, -2.5) {\small Gradients};
        \node[prmColor, anchor=west] at ($(params) - (1.15, -0.55)$) {\small Loihi Parameters};
        \node[gray    , anchor=west] at ($(weights) + (0.4, 0.2)$) {\small Shadow Weight};

        \draw[thick, fwdColor, fill=white, fill opacity=0.8] ($(weightedSpikes.west) - (0, \h)$) rectangle ($(weightedSpikes.east) + (0, \h)$);
        \draw[thick, fwdColor, fill=white, fill opacity=0.8]
            ($(spikeResponse.west) - (0.2, \h)$) rectangle 
            ($(spike.east)         + (0.2, \h)$);
        \draw[thick, fwdColor!50, fill=white, fill opacity=0.8] 
            ($(spikeResponse.west) - (0, \h) + (0, \dh)$) rectangle 
            ($(spikeResponse.east) + (0, \h) - (0, \dh)$);
        \draw[thick, fwdColor!50, fill=white, fill opacity=0.8] 
            ($(spike.west) - (0, \h) + (0, \dh)$) rectangle 
            ($(spike.east) + (0, \h) - (0, \dh)$);
 
        \draw[thick, black!40, ->, dashed] 
            (gradW) -|
            ($(weights) - (1, 0.5)$) -- ($(weights) + (\r, 0.5*\r)$); 
 
    \end{scope}
\end{tikzpicture}
\caption{Computational blocks for offline pre-training of SNN for Loihi using SLAYER. The SNN is modeled with a functional Loihi simulator and the normalized post-synaptic response is used for temporal credit assignment. Since Loihi uses integer weights full precision shadow weights are quantized and used during the forward inference phase.} 
\label{fig:slayerLoihi}
\end{figure}
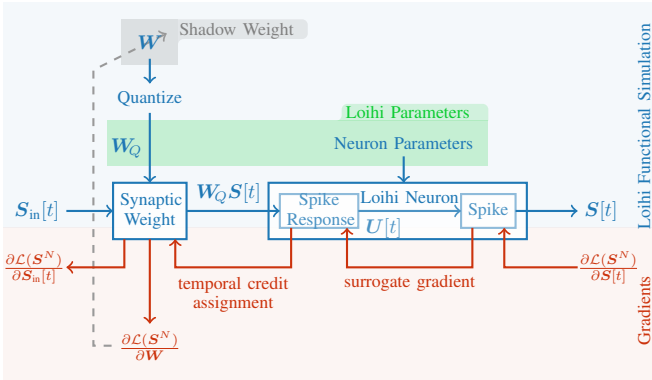

Assuming a global cost function $\mathcal{L}(S^{N})$ defined on the spikes $S^N$ of the top layer and targets ${Y}$, the gradients with respect to the weights in layer $l$ are:
\begin{equation}\label{eq:loss}
\nabla_{W_{ij}} \mathcal{L} (S^{N}) =
\frac{\partial \mathcal{L} (S^{N})}{\partial S_{i}} 
\frac{\partial S_i          }{\partial U_{i}} 
\frac{\partial U_i          }{\partial W_{ij}}.
\end{equation}
We discuss below the three factor above. For didactic reasons, we proceed first with the middle term, then the first term, then the third.
The middle term is the derivative of the activation function $\Theta$ of the spiking neuron which is non-differentiable.
As discussed earlier, the \ac{SG} approach consists in using a smooth surrogate function in place of the non-differentiable step function, such as the boxcar function \cite{Neftci_etal17_evenranda,Kaiser_etal20_synaplas}. 
The first term on the right-hand side describes how the loss changes as the spiking states in the network, $S_{i}$, change.
If the loss function is the mean-squared error and the network consists of only one layer, the first term becomes the task error $(Y_i-S_i^{N})$.
Computing $\frac{\partial \mathcal{L} (S^{N})}{\partial S_{i}}$ for hidden layers is non-trivial and equivalent to solving a spatiotemporal credit assignment problem. 
Two methods exist to solve this problem: (1) it can be computed \emph{offline} using gradient backpropagation on the time-unfolded graph (\emph{i.e.} \ac{BPTT}), or \emph{online} by using local loss functions \cite{Kaiser_etal20_synaplas}.
This work uses a combination of offline and online \ac{SNN} learning, namely SLAYER and \ac{SOEL} for pre-training hidden layers and online three-factor rules for learning in output layers, respectively. 
In the following paragraphs, we provide further detail about these two learning methods. 

\subsubsection{\acf{SOEL} Online training for Loihi}
Online training on physical substrate requires all the information necessary for computing the gradient to be available at the synaptic plasticity processor.
The first two terms of \refeq{eq:loss} discussed above are errors and postsynaptic states.
The last term in \refeq{eq:loss} can be computed from Eq. (\ref{eq:clif}--\ref{eq:pq}) (or \refeq{eq:lif_equations} in the discrete case).
The derivative of the reset term introduces the full history of the spiking neuron, which cannot be computed locally in time.
However, in low firing rate regimes, the error in omitting this term in the gradient calculation is small. By omitting the reset process, the third term becomes simply the trace $P_{j}$.
Finally, we are left with the following three factors:

\begin{equation}\label{eq:eos_cont}
\nabla_{W_{ij}} \mathcal{L} (S) =
- (Y_i-S_i) 
\sigma'(U_i) 
P_j.
\end{equation}
Provided that pre-synaptic traces, membrane potential and errors are available at the synapse, learning can be performed locally as a synaptic plasticity rule.
In computational neurosciences, rules of this type are referred to as three-factor rules \cite{Gerstner_etal18_eligtrac}. 
Three-factor rules are consistent with biological synapse dynamics and constitute a normative theory of learning in the brain.

\refeq{eq:eos_cont} prescribes updates at every timestep. 
While this is consistent with biological dynamics, it is not efficient in hardware.
Updates can instead be made when the error $(Y_i-S_i^N)$ crosses a threshold, thus forming a binary ``error event'' \cite{Payvand_etal20_errothre}.
This is reminiscent of \ac{STDP}, where updates are triggered when pre-synaptic or post-synaptic neurons events occur \cite{Bi_Poo98_synamodi}.
Here, updates are instead triggered by error events. 
Error-triggered learning allows the conditional activation of the plasticity operations, which can drastically reduce the footprint of online learning.
Recent work showed that the number of updates can be reduced by 20 fold for a small loss in accuracy \cite{Payvand_etal20_errothre}.
Using a piecewise \ac{SG} function, $B_i =\sigma'(U_i)$ becomes a box function where $B_i \in \{0,1\}$. Then, the \ac{SOEL} rule can be written in the following compact, three-factor form:
\begin{equation}\label{eq:seol}
\nabla_{W_{ij}} \mathcal{L} (S^{N}) 
\propto
- E_i
B_i 
P_j.
\end{equation}
where $E_i$ is a integer error event for neuron $i$.

\subsubsection{SLAYER Offline Training for Loihi}
\label{sec:slayerLoihi}


SLAYER is a gradient computation method for training deep SNNs directly in the spiking domain~\cite{Shrestha_Orchard18_slayspik}. It treats the inputs and outputs of the \ac{SNN} as temporal signals and backpropagates the error at the output layer accordingly. There are two basic guiding principles in SLAYER: Temporal error credit assignment, and the surrogate gradient.
%
%
Temporal credit assignment is done by unfolding the temporal dynamics in time and backpropagating through the unfolded graph.
Further, it is typical for the normalized post-synaptic response to decay to practically zero after some time. 
Therefore, it is sufficient in practice to apply temporal error credit assignment only up to a finite point in history.
SLAYER uses a proxy function as an approximation of spike function derivative, similar to the surrogate gradient learning described in Section~\ref{sec:surrogateGradient}. 
These principles form the essential link in the computational graph used to calculate the gradients of the weights of the \ac{SNN} and train it using standard deep learning optimization methods.

SLAYER PyTorch\footnote{
 \url{https://github.com/bamsumit/slayerPytorch}}  also supports training an SNN with the CUBA leaky integrate and fire neuron model compatible with the Loihi chip. 
For one-to-one mapping of the trained network in Loihi hardware, the \ac{SNN} is modeled with a functional Loihi simulator and the normalized post-synaptic response is used for temporal error credit assignment during backpropagation. 
In addition, since Loihi only supports integer weights, a strategy of full precision shadow weights~\cite{hubara2017quantized, Courbariaux2015} is used, which are quantized during the forward inference phase only. 
The computational blocks for SLAYER-Loihi training are shown in ~\reffig{fig:slayerLoihi}.

\subsection{Intel Loihi}
The Intel Loihi is a neuromorphic processor that integrates a wide range of features such as hierarchical connectivity, dendritic compartments, synaptic delays, and programmable synaptic learning rules \cite{Davies_etal18_loihneur}.
Each Loihi chip is composed of a many core mesh comprising of 128 neuromorphic cores with each core implementing 1024 primitive spiking neural units, three embedded x86 processor cores, with an asynchronous network-on-chip (NoC) for between core communication. Loihi offers a variety of local information for programmable synaptic learning processes such as spike traces with configurable time constants that can have different time constants.  
\subsubsection{Plasticity Processor}
Synaptic weights can be updated via a learning rule expressed as a finite-difference equation with respect to a synaptic state variable that follows a sum-of-products form as follows \cite{Davies_etal18_loihneur}:
\begin{equation}\label{eq:loihi_rule}
    W_{ij}[t+1] = W_{ij}[t]+\sum_k C_{k}\prod_l F_{kl}[t],
\end{equation}
where $W_{ij}$ is the synaptic weight variable defined for the destination-source neuron pair being updated; $C_k$ is a scaling constant; and $F_{kl}[t]$ may be programmed to represent various state variables, including pre-synaptic spikes or traces, post-synaptic spikes or traces, where traces are represented as first-order linear filters. 
The weights are stochastically rounded according to the programmed weight precision. Traces are stochastically rounded to 7-bits of  precision.

\section{Methods}
\begin{figure*}
    \includegraphics[width=1.0\textwidth]{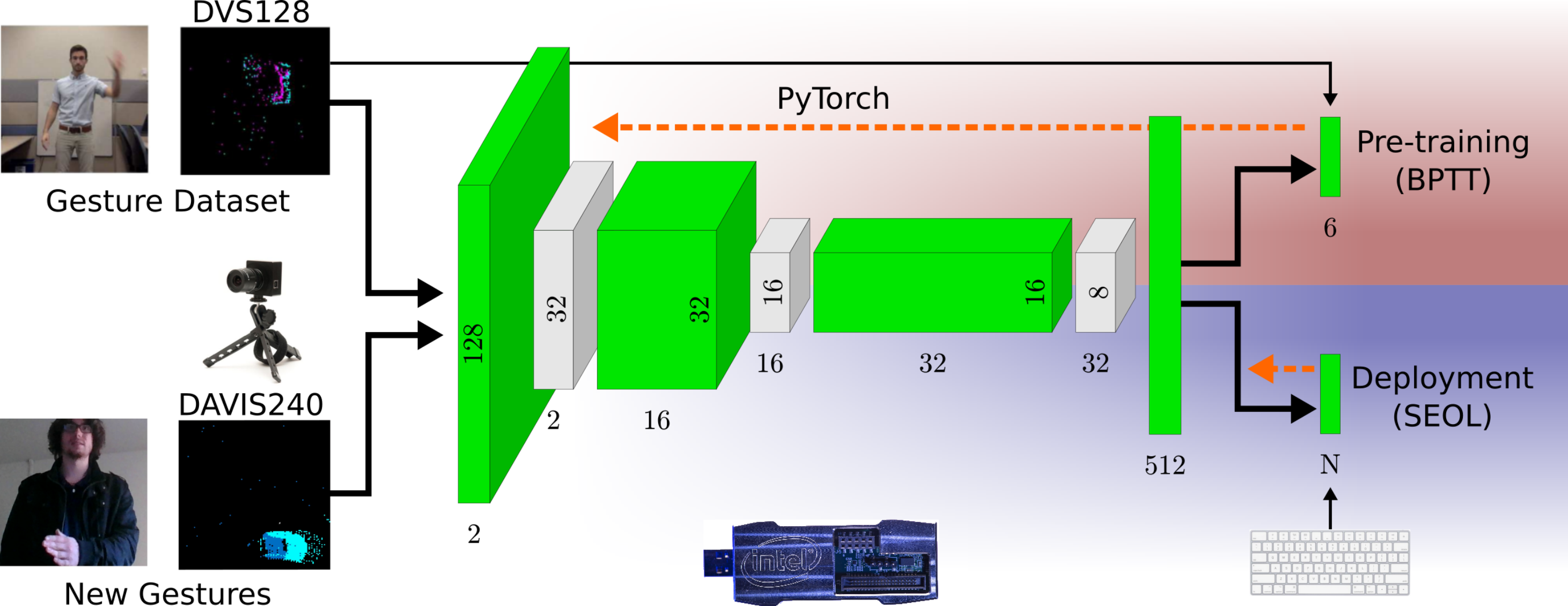}
    \centering
    \caption{Experimental setup. During a pre-training phase, the Loihi compatible convolutional network is trained on a computer using an event-based gestures dataset, the functional simulator, and SLAYER/\ac{BPTT}. In this work, the pre-training dataset consisted of the IBM DVS Gestures dataset recorded using a DVS128 camera. The entire network along with quantized parameters of the functional simulation are then transferred on to the Loihi cores. During deployment, new gestures recorded using a DAVIS are streamed to an Intel Kapoho Bay. Few-shot learning is performed on the final layer using on-chip \ac{SOEL}. The deployed network, including inference and training dynamics are performed on the Loihi chips. Dashed orange arrows indicate the extent of the spatial credit assignment, and thus which layers are trained in each of the two phases.}
    \label{fig:learning}
\end{figure*}
We present a system for online learning of gestures from DVS data using only a few shots.
Our workflow consist of a pre-training phase, followed by a deployment phase. 
The pre-training phase uses SLAYER and its functional Loihi simulator to train a Loihi compatible convolutional network on a GPU.
For our targeted human gesture recognition application, we use the event-based IBM DVS Gestures dataset to pre-train this network. 
The trained network and quantized parameters are then transferred to the Loihi cores. During deployment on Loihi, few-shot learning of new gestures is performed on the top layer on-chip with \ac{SOEL}. The system is shown in figure \ref{fig:learning} and its components are detailed in the following subsections. 

\subsection{Dataset and gesture sampling} 
Visual input to the model was recorded with either a DVS 128 in the case of the IBM DVSGesture\footnote{The DVS Gesture dataset is used under a Creative Commons Attribution 4.0 license.} dataset~\cite{Amir_etal17_lowpowe} or with a DAVIS 240C~\cite{Brandli_etal14_240180} in the case of real-world gestures.
The network was pre-trained data recorded with a DVS 128, which has a smaller resolution compared to the more recent DAVIS 240C.
During experiments involving input live-streamed from a DAVIS 240C to an Intel Kapoho Bay, data was scaled down to the same dimensions as the DVS 128 before being input into the network.
Data was taken by one subject under three different lighting conditions, natural light from the sun, incandescent light, and fluorescent light which is shown in \reffig{fig:demo}. 
%

\subsection{Neural Network Model and Offline pre-training using SLAYER for Loihi} 
\begin{table}[!ht]
\label{tab:arch}
\begin{center}
\caption{Network architecture.}
\def\x{$\times$}
\begin{tabular}{|c|c|c||c|} \hline
Layer & Kernel & Output    & Training Method\\ \hline
input &      & 128\x128\x2 & DVS128/DAVIS240C (Sensor)\\\cline{4-4}
1     & 4a     & 32\x32\x2   & \multirow{6}{*}{SLAYER (BPTT)}\\
2     & 16c5z   & 32\x32\x16  &\\
3     & 2a     & 16\x16\x16  &\\
4     & 32c3z   & 16\x16\x32  &\\
5     & 2a     & 8\x8\x32    &\\
6     & -      & 512       &\\ \cline{4-4}
output   & -      & N        & SOEL\\ \hline
\end{tabular}
\end{center}
\scriptsize{Notation: \verb~Ya~ represents \verb~YxY~ sum pooling, \verb$XcYz$ represents \verb$X$ convolution filters (\verb~YxY~) with zero padding. $N$ is the number of classes, which is task-dependent.}
\end{table} 
We trained a spiking CNN using SLAYER for Loihi (c.f. Section~\ref{sec:slayerLoihi}) on the DVS Gesture dataset~\cite{Amir_etal17_lowpowe}.
It has eleven output gestures, out of which six (the even classes) were used for offline training using SLAYER.
The input is a $128 \times 128$ spatial event with two polarities (\textsc{on} and \textsc{off}).
The spiking CNN architecture shown in table I consisted of 7 layers.
The input spikes are OR'ed in 1~ms time bins and then fed to the network.
$N$ refers to the number of output classes, which depend on the experimental conditions.

The threshold for all the neurons were set to $80\times 2^6$ and the current decay and voltage decay were set to $1024$ (time constant of 32ms) and $128$ (time constant of 4ms) respectively. The weights of the network were trained to be in the set $\{-256, -254, \cdots, 254\}$ i.e. 8 bit signed weights with step of 2.

The network was pre-trained for $2000$ epochs. For better generalization performance, the input was augmented during training: x-y jitter of up to $8$ pixels, rotation jitter of up to $10^\circ$, and random sampling of 1450ms spike sequence. The Nadam\cite{dozat2016incorporating} optimizer was used with a learning rate of $0.003$ and default $\beta = (0.9, 0.999)$.
The network without the final fully connected output layer (layers $1$--$6$), is the feature extraction network which is subsequently used in our on-chip learning experiments described below.

\subsection{Online few-shot learning using \acf{SOEL} plasticity for Loihi } 
\ac{SOEL} requires the pre-synaptic trace $P$ to be a second-order linear filter.
Second-order kernels can be implemented as a subtraction of two first-order kernels \cite{Gerstner_etal14_neurdyna}. 
This subtraction is enabled by the sum-of-products formulation of the plasticity rule (\ref{eq:loihi_rule}).
The error, $err_i$, is computed with the post-synaptic neuron using the following:
\begin{align}  \label{eq:error}
    err_i[t] &= Y - \bar{S}_i[t]   
\end{align}
where $Y$ is the target, $\bar{S}_i = \sum_{t-T}^T S_i[t]$ is defined here as the number of post-synaptic spikes by neuron $i$ in the previous $T$ timesteps. $T$ is a constant number of timesteps that is a fraction of the total presentation time of the sample. 
This number determines the rate at which errors are computed.
Using a spike-count instead of spike states is an approximation because the update will be subsequently made using the states in the final timestep, \emph{i.e.} $P_j[t]$.
However, for $T$ smaller than the neuron and synaptic time constants, $P_j[t]$ will not vary much during this time window, and the approximation will remain close to the exact case.

Since the post-synaptic trace is not necessary for the \ac{SG} rule, \ac{SOEL} writes the error on the same register used for the post-synaptic trace. This enables the error value to be available in the plasticity processor for learning.
On the chip, post-traces can only be positive but errors can be both positive and negative.
This problem is solved by offsetting the weight updates with a constant term $C$.

\begin{equation}\label{eq:eos_err}
    E_i[t] = \begin{cases} 
                C + {err}_i[t],\text{ if $err_i > \theta$ or $< -\theta$}\\
                C, \text{otherwise}
                \end{cases}
\end{equation}
 where $\theta$ is an error threshold. 

Intuitively, \ac{SOEL} can be interpreted as follows.
If ${err}$ is higher than $\theta$, meaning the neuron is spiking at too high a frequency, then there is a positive error and the weight of the synapse will be penalized. 
Conversely if ${err}$ is below $-\theta$ then the weight of the synapse weight will be increased.
The term $\sigma'$ (the ``second factor'' in \refeq{eq:eos_cont}) cannot be implemented directly on the Loihi because the membrane state is not available at the plasticity processor.
Since only the final layer $N$ is trained, setting this term to 1 regardless of the membrane value only has the effect of continued learning even after the neuron output saturates in either direction.
This strategy, referred to straight-through estimator in the machine learning field, has the disadvantage of yielding biased estimated of the gradients, but the advantage of faster learning since every error leads to an update. 
Since our goal is to perform fast, one-shot learning, \ac{SOEL} implemented here uses a straight-through estimator.
The full learning rule can then be expressed as:
\begin{equation}
    W_{ij}= W_{ij} + \eta (E_i-C) P_,
    \label{eq:LR}
\end{equation}

\begin{algorithm}
\SetAlgoLined
\KwResult{Error-triggered Synaptic Plasticity}
$\theta$ = 0\;

\If{neuron i is learning}{
$\bar{S}_i \leftarrow \bar{S}_i + S_i$;\\
${err}_i \leftarrow Y - \bar{S}_i$\;
\eIf{${err}_i > \theta$ or ${err}_i < -\theta$}{
$E_i \leftarrow C + {err}_i$\\
$W_{ij} \leftarrow W_{ij} - \eta (E_i-C) P_j$;\\
increase $\theta$\ by constant;\\
$\bar{S}_i \leftarrow 0$;\\
}
{decrease $\theta$ by constant; 
}}
\label{alg:SOEL}
\caption{SOEL}
\end{algorithm} 
where $W_{ij}$ is the synapse from pre-synaptic neuron j to post-synaptic neuron i, $\eta$ is the learning rate, $E_i$ is the error, and $P_j$ is the pre-synaptic trace.
The learning rule can be implemented in Intel Loihi as:
\begin{equation}
    \begin{split}
    X^1_j[t+1] &= \alpha^1 X^1_j[t] + S^1_j,\\
    X^2_j[t+1] &= \alpha^2 X^2_j[t] + S^2_j,\\
    Y_i[t] &= E_i[t],\\
    \Delta W_{ij} &= \eta (X^2_j[t]-X^1_j[t])(Y_i[t] - C) .
    \end{split}
\end{equation}
Here, $X^2$ and $X^1$ are pre-synaptic trace variables available in the Loihi whose subtraction in the third equation yields the second order kernel equivalent to $P_j$ in \refeq{eq:pq}. 
\begin{equation}
P_j [t] \propto (X^2_j[t]-X^1_j[t]).
\end{equation}
A Loihi Lakemont core computes the spike count $\bar{S}$ and evaluates $err_i$ at regular intervals $T$. If the error exceeds the threshold $\theta$, the post-synaptic trace value in the plasticity processor, $Y$, is written with the error $E_i$ and a plasticity operation is initiated. 
As in \refeq{eq:eos_err}, $C$ is a constant bias term to account for negative error because traces cannot be negative. 
\begin{figure}[ht!]
    \includegraphics[width=.37\textwidth]{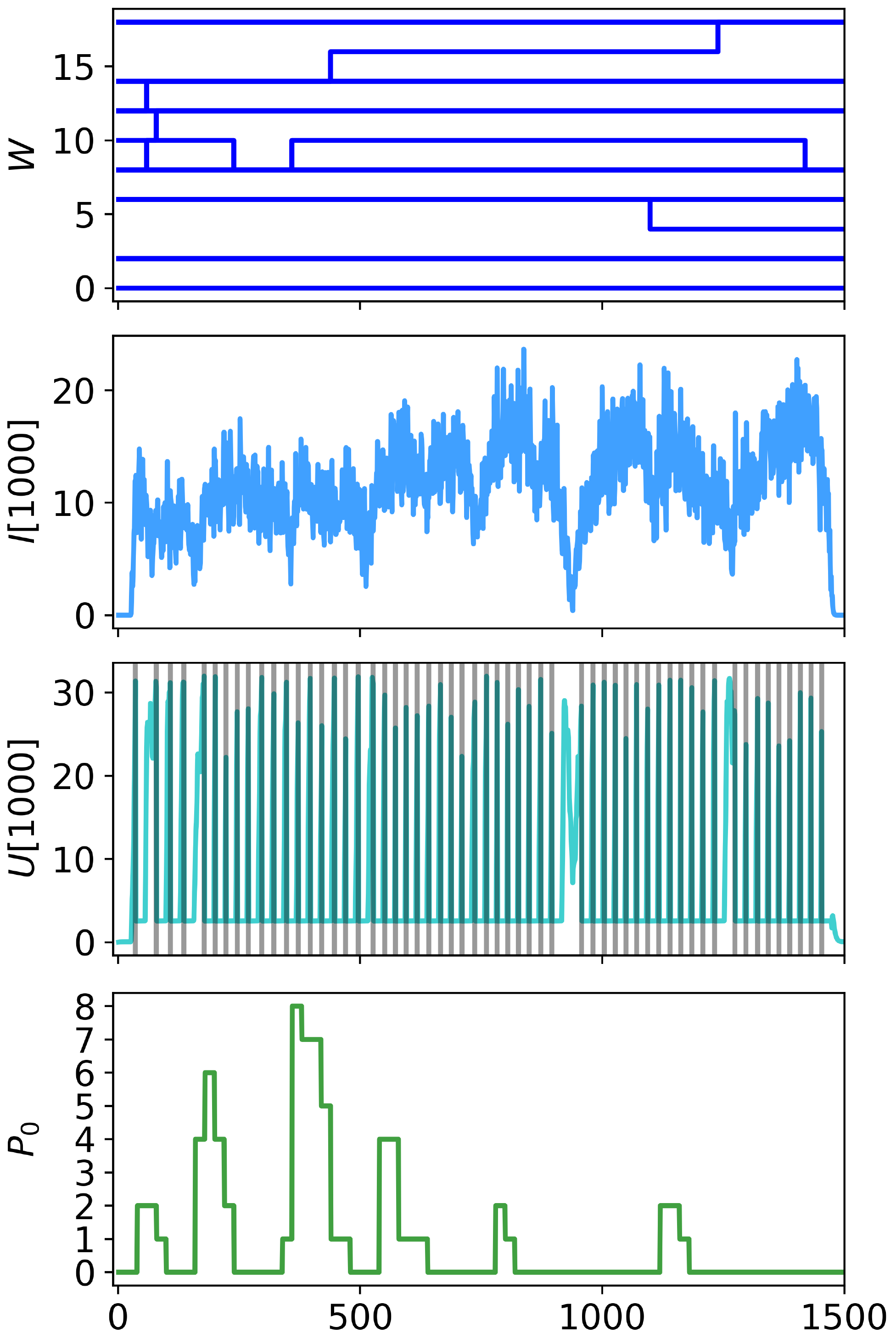}
    \centering
    \caption{Dynamics of one learning neuron when learning a new gesture. Only a subset of the synaptic weights W are shown. The weights only change when $P_0$ is non-zero during a learning epoch. The current I, and membrane potential U of the learning neuron are shown over the duration of the sample. Spikes are shown as grey vertical lines overlaying the membrane potential plot. 
    } 
    \label{fig:dynamics}
\end{figure}

\subsection{System Specifications For Measurement \footnote{All performance results are based on testing as of June 2020 and may not reflect all publicly available security updates. No product can be absolutely secure.}}
SNN offline pre-training was performed with Ubuntu 16.04.6, SLAYER PyTorch commit id 598fc44, and PyTorch 1.4.0. The machine consists of an Intel Xeon E5-2630 CPU with 128GB RAM and Nvidia GeForce RTX 2080Ti GPU.

Loihi time and energy measurements were made using Ubuntu 16.04.6 with Nx SDK 0.95 and a Nahuku 32 board running on the Intel Neuromorphic Research Community (INRC) cloud. The machine consists of an Intel Xeon E5-2650 CPU with 4GB RAM. 

Live gesture learning used a Kapoho Bay Loihi system connected to an IniVation DAVIS240C sensor. The host machine was an Intel Core i7-7700HQ CPU with 16GB of RAM running Ubuntu 16.04.6 and Nx SDK 0.95.

\section{Experiments}
We used SOEL to train and test the last layer of the neural network pre-trained with SLAYER on 6 of the 11 gestures from the DVSGesture dataset, training the last layer with only a few-shots of the remaining 5 gestures of the dataset for a few-shot 6+5 way gesture classification task.
The 6 refers to the 6 gestures the network is pre-trained to classify using SLAYER, and the 5 refers to the 5 new gestures we are training the last layer of models on using only a few-shots.
``Train'' refers to classification accuracy on training samples using saved weights with plasticity disabled.
``Test'' is the classification accuracy on the samples held out of the training procedure.
Models were trained on one, five, or twenty shots of data and then tested on 100 held out samples.
The results are obtained from performing a 5-fold cross-validation.
A gesture is considered correctly classified if the desired neurons spike frequency is highest during the presentation time.
We compare the accuracy of the model using SOEL to two other models, one whose last layer is trained using vanilla SGD used in \cite{Stewart_etal20_on-cfew-} and another whose last layer is trained using SLAYER. 
Each model was trained on samples from the DVSGesture test dataset, and tested on samples from the test dataset not seen during training. Table \ref{tab:dvs_table5-way} shows the accuracy comparisons of the different models trained on the few-shot 6+5-way gesture classification task. 
The results show the SOEL trained network is overall better than the vanilla SGD method from \cite{Stewart_etal20_on-cfew-}, achieving on average significantly better results at test time after seeing only one shot of training data, and better generalization.
However while SOEL does better at training time on 1 shot experiments than the pure SLAYER network, SLAYER is better at generalizing than SOEL. 
This could be due to the SOEL model tending to over-fit on samples presented and the straight-through estimator.
For similar reasons, we speculate that the accuracy of the SOEL model is more variable than SLAYER. When overfitting, samples that deviate too much from those samples will be more likely to be classified incorrectly, but all experiments show SOEL to be significantly better than our previous implementation \cite{Stewart_etal20_on-cfew-}. 
We also compare the time taken and energy consumption needed for SOEL and \cite{Stewart_etal20_on-cfew-} to train each gesture shown in table \ref{tab:time-energy}. The results indicate that SOEL uses more energy but takes less time to train and achieves higher accuracy than \cite{Stewart_etal20_on-cfew-}. Because the Intel Kapoho Bay does not support energy probing, energy and time measurements were taken with an Intel Nahuku board consisting of 32 Loihi chips. 

\begin{table}[!t]
\centering
\caption{\label{tab:dvs_table5-way}6+5-way few-shot classification on the DvsGesture dataset 
}
\begin{tabular}{|c|c|c|c|c|}
\hline
Dataset                         & Learning Method                                   & Shots       & Train           & Test            \\ \hhline{|=|=|=|=|=|}
\multirow{9}{*}{DVSGesture    } & \multirow{3}{*}{SOEL}            & \textbf{1}  & \textbf{96$\pm$4\%}   & \textbf{64.7$\pm$4.6\%}   \\ \cline{3-5}
                                &                                                   & \textbf{5}  & \textbf{88$\pm$5.2\%}   & \textbf{65.1$\pm$5.1\%} \\ \cline{3-5}
                                &                                                   & \textbf{20} & \textbf{87.7$\pm$2.3\%} & \textbf{80.2$\pm$4.3\%} \\ \cline{2-5}
                                & \multirow{3}{*}{SGD \cite{Stewart_etal20_on-cfew-}}                   & 1           & 40\%            & 40\%            \\ \cline{3-5}
                                &                                                   & 5           & 60\%            & 43.3\%          \\ \cline{3-5}
                                &                                                   & 20          & 73.5\%            & 56.2\%          \\ \cline{2-5}
                                & \multirow{3}{*}{SLAYER}                   & 1           & 78.5$\pm$2.6\%            & 83.5$\pm$2.32\%            \\ \cline{3-5}
                                &                                                   & 5           & 95.9$\pm$1\%            & 83.5$\pm$2.9\%          \\ \cline{3-5}
                                &                                                   & 20          & 99.7$\pm$.3\%            &  91.2$\pm$1.9\%            \\ \hline
\end{tabular}
\end{table}

\begin{table}[!t]
\centering
\caption{\label{tab:time-energy}Comparison of time and energy taken for learning one gesture
}
\begin{tabular}{|l|l|l|r|}
\hline
Measurement          & SOEL   & SGD \cite{Stewart_etal20_on-cfew-}    & \multicolumn{1}{l|}{\% Diff} \\ \hhline{|=|=|=|=|}
Learning Time (s)   &  .037   & .31   & -87.71\%                       \\ \hline
Learning Energy (mJ) &  167.58  & 37.04   & 358.46\%                         \\ \hline
Learning Power (mW) & 6.18 &	11.48	& -46.17\%                      \\ \hline
Total Time (s)      & 1.04 & 1.21 &  -14.23\%                      \\ \hline
Total Energy (mJ)    & 481.98 & 323.2 & 49.13\%                      \\ \hline
Total Power (mW) &  511.65	& 391.07	& 30.83\%      \\ \hline
\end{tabular}
\end{table}

\subsection{Real-World Gesture Learning}
\begin{figure}[ht!]
\centering
    \includegraphics[width=.3\textwidth]{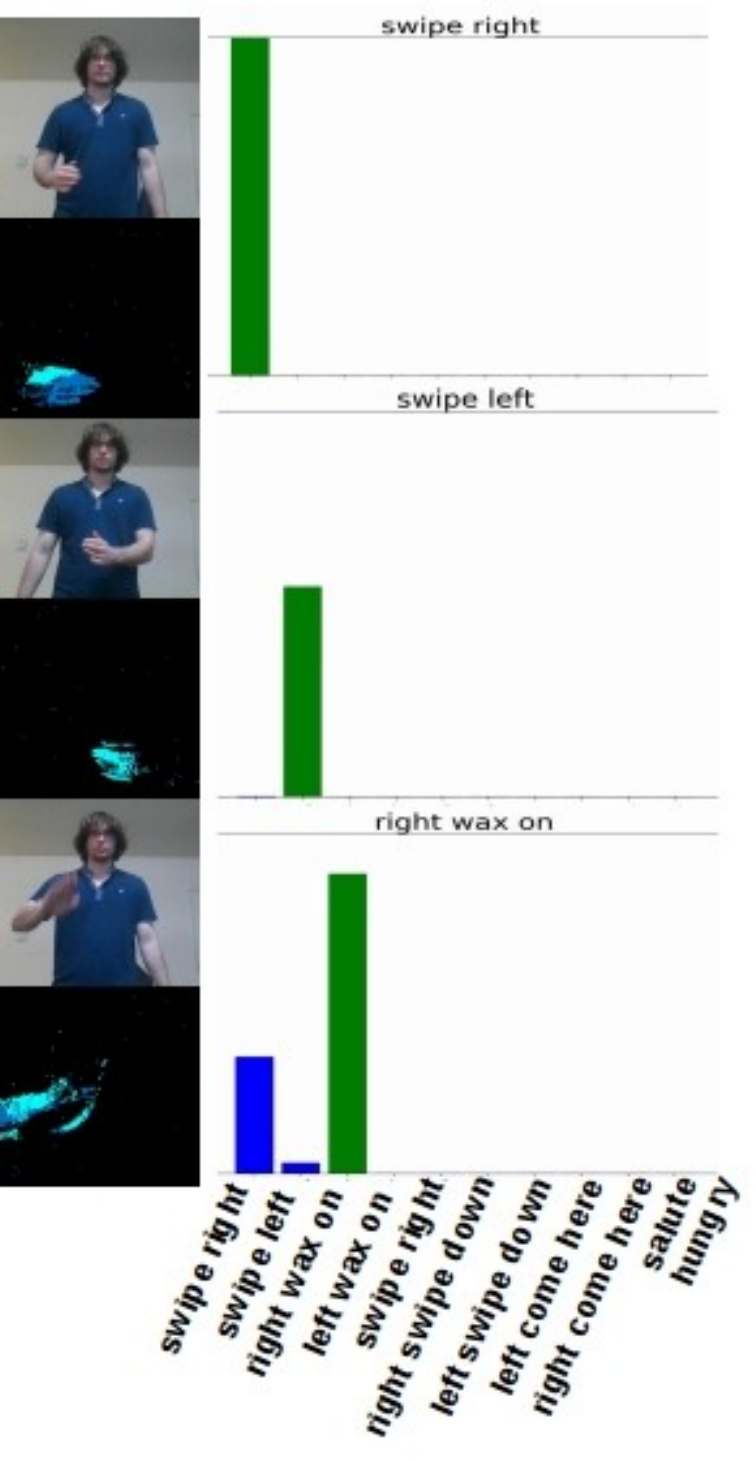}
    \caption{Rapid online learning of gestures using data streamed from a DAVIS240C to an Intel Kapoho Bay. The upper part of the figure shows a person performing a gesture in front of a DAVIS240C, and the corresponding DAVIS240C output events shown in blue. The histogram shows the spiking frequency of each neurons response to the presented gesture after learning. After only a single one second presentation of each gesture the network can correctly classify the gestures it trained on.}
    \label{fig:demo}
\end{figure}

In addition to the few-shot 6+5 way classification we also tested SOEL in a real-world gesture learning and recognition setting. To demonstrate rapid online gesture learning in a real-world setting we streamed gesture data in real time from a DAVIS 240C sensor connected to an Intel Kapoho Bay. For the experiment we tested one subject in a single lighting condition where the subject was under fluorescent light. The neural network model on the Kapoho Bay was pre-trained on all 11 gestures of the DVSGesture dataset using SLAYER, but the last layer is reset, made plastic, and trained using SOEL. The task was to train the network to classify 10 predetermined gestures outside of the DVSGesture dataset using as few shots as possible. Figure \ref{fig:demo} shows an example of the learning and inference of the gestures. 
After being shown a gesture for a one second presentation, the network is able to classify other samples of the gesture. 
Additionally, training other gestures does not interfere with the networks ability to classify previously learned gestures. However, performance can degrade if the learned gestures spatially overlap because unique gestures within the same space may be seen as the same gesture.

The results of which some are shown in figure \ref{fig:demo} demonstrate the capability of the SOEL learning rule to perform rapid few-shot learning on a neuromorphic processor from real-world data. A link to a video showing a live demonstration of the rapid learning of 10 new gestures is added as supplemental information.

\section{Discussion and Future Work}
We presented SOEL, a new surrogate gradient based learning algorithm for few-shot online learning on an Intel Loihi neuromorphic processor using gesture recognition as a case study. To accomplish this we first pre-trained an Intel Loihi compatible \ac{SNN} on a GPU using the current state-of-the-art SLAYER method, and then deployed the network on an Intel Kapoho Bay and retrained the last layer on few-shots of data using SOEL. 
We found that like \acp{ANN}, using a pre-trained network for transfer learning with \acp{SNN} significantly boosts few-shot learning accuracy.
While we have achieved real-time online gesture learning using SOEL, there are limitations to our method. Currently, SOEL only supports training the last layer of the network. Being a local learning rule, SOEL only has information from pre- and post-synaptic neurons within its layer. Therefore training other layers will incur the spatial credit assignment because the neurons will not have a direct target to train on outside of the last layer.
Consequently, if the signal is not separable in the penultimate layer then the last layer cannot learn. 
This can be potentially solved using layer wise local loss functions \cite{Kaiser_etal20_synaplas} and is beyond the scope of this article.
Another limitation stems from the approximations made with the SOEL algorithm. First, the algorithm assumes that states do not change across the time window in which the error is calculated. This is beneficial to speed up training and can be adjusted to match the error dynamics. Second, due to limitations of the plasticity processor, the second term of the three factor rule cannot be implemented exactly and is instead ignored (set to one). These two approximations are likely to reduce the accuracy of the final result.
The few-shot learning experiments using SOEL on gesture data with the Intel Loihi neuromorphic processor are slightly worse compared to training the last layer using GPU SLAYER.
This discrepancy is expected since SOEL yields biased estimates of the gradients.
The bias in the estimates is caused mainly by the straight-through estimator, and the approximate spike count loss which is computing using the neural states of the last time step. Furthermore, the discretization of neural and synaptic states, and limited range of effective learning rates further widen the gap between GPU simulations and Loihi simulations.
However, in the regime of interest, \emph{e.g.} between one shot and five shots, the discrepancy remains acceptable. Furthermore, they are a major improvement from our previous work. 
Unlike vanilla SGD, which learns at every timestep, SOEL only learns when there is sufficient error to trigger learning. 
This error-triggered learning helps prevent weight saturation and catastrophic forgetting leading to increased accuracy. 
However the increased accuracy comes with an increase in power consumption when compared to vanilla SGD. 
We speculate that the power consumption for gesture recognition using SLAYER with a GPU is at least an order of magnitude higher than using SOEL with the Intel Loihi. 
Additionally we also showed SOEL is capable of few-shot learning from real world data. 
These experiments also showed SOEL was able to adapt to the differences of data taken from both a DAVIS 240 and a DVS 128 and was able to learn using data from both.

\section*{Acknowledgements}
The preliminary experiments of this research were conducted at the Telluride Neuromorphic Cognition Engineering workshop, years 2018 and 2019 (all authors). This research was supported by the Intel Corporation (KS, EN), the National Science Foundation under grant 1652159 (EN), and partially supported by Programmatic grant no. A1687b0033 from the Singapore government’s Research, Innovation and Enterprise 2020 plan (Advanced Manufacturing and Engineering domain) (SBS).
\vfill
\pagebreak
\bibliographystyle{IEEEtran}
\bibliography{biblio_unique_alt,bib}

\begin{IEEEbiography}[{\includegraphics[width=25mm,height=32mm,clip,keepaspectratio]{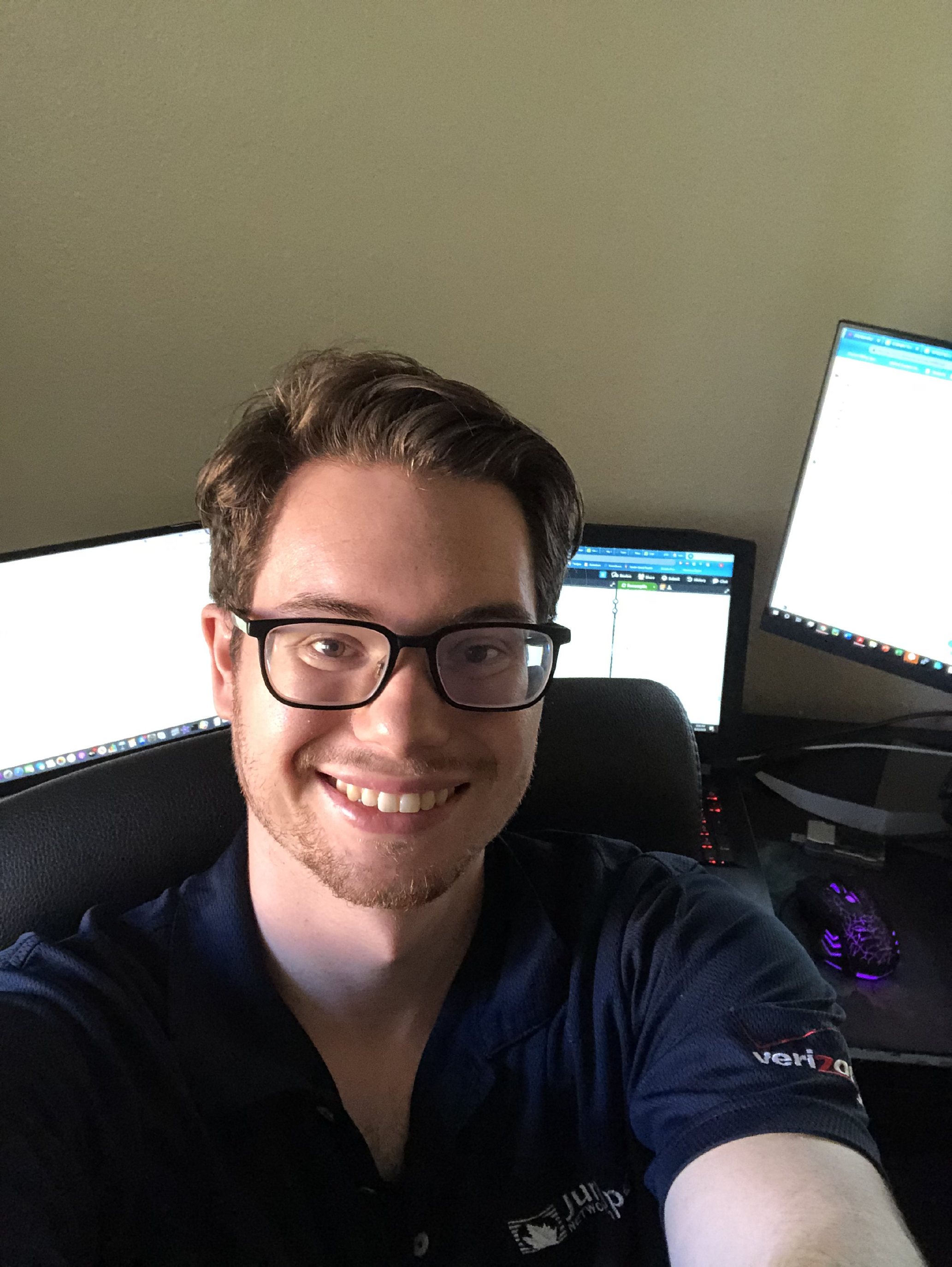}}]{Kenneth Stewart}
is a Ph.D. student at the University of California Irvine. His current research focuses on developing learning algorithms for neuromorphic hardware and their application to areas such as computer vision and robotics. His research interests include neuromorphic computing, online learning, robotics, artificial intelligence, and applications thereof.
\end{IEEEbiography}

\begin{IEEEbiography}[{\includegraphics[width=25mm,height=32mm,clip,keepaspectratio]{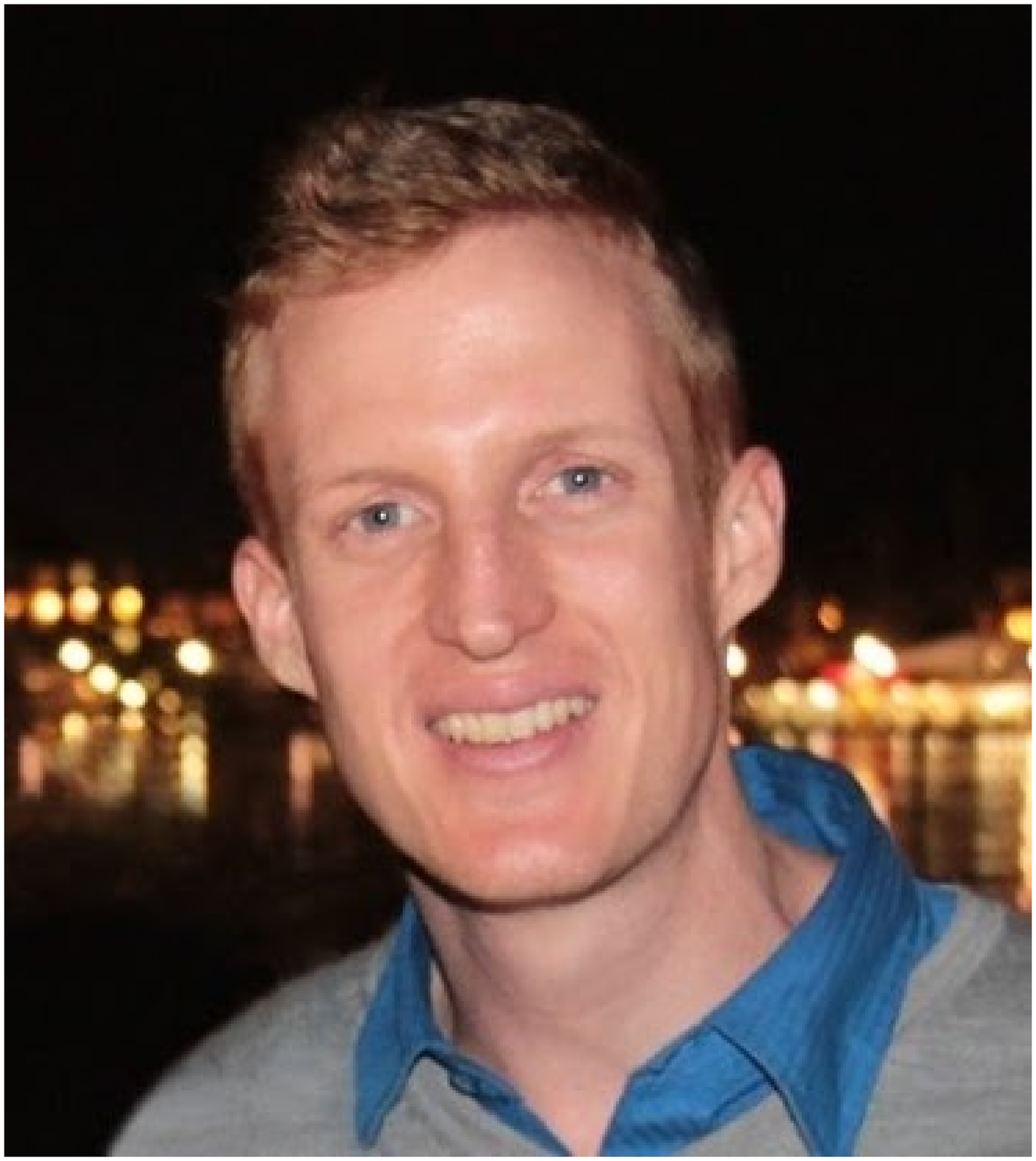}}]{Garrick Orchard}
received his Ph.D. from Johns Hopkins University in 2012 before joining the newly formed Singapore Institute for Neurotechnology (SINAPSE) at the National University of Singapore as a research scientist. In 2015 he was awarded the Temasek Research Fellowship from the Singapore Ministry of Defence and in 2019 he joined Intel’s Neuromorphic Computing Lab as a senior researcher focusing on sensing and perception. 
\end{IEEEbiography}

\begin{IEEEbiography}[{\includegraphics[width=1in,height=1.25in,clip,keepaspectratio]{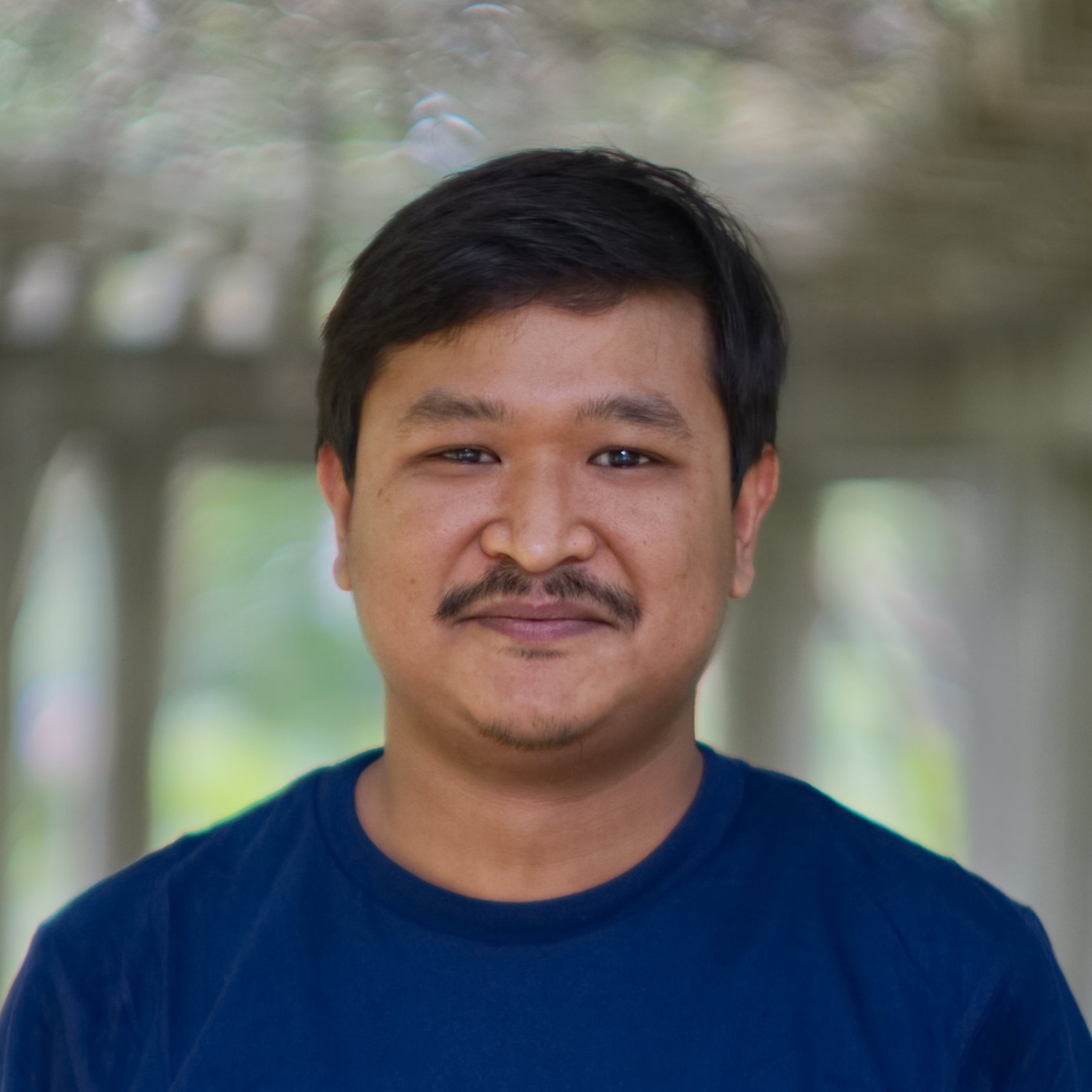}}]{Sumit Bam Shrestha} 
received his Ph.D.  from School of Electrical and Electronic Engineering, at Nanyang Technological University, Singapore, under SINGA scholarship. Currently, he is a Research Scientist at the Institute for Infocomm Research~(I2R) at the Agency of Science Technology and Research~(A*STAR) Singapore where he co-leads the algorithm development for Neuromorphic Computing Programme.
His research is mainly focused on Deep Spiking Neural Networks and also includes Neuromorphic Computing, Neuromorphic Vision, Neural Networks, and Machine Learning.
\end{IEEEbiography}

\begin{IEEEbiography}[{\includegraphics[width=1in,height=1.25in,clip,keepaspectratio]{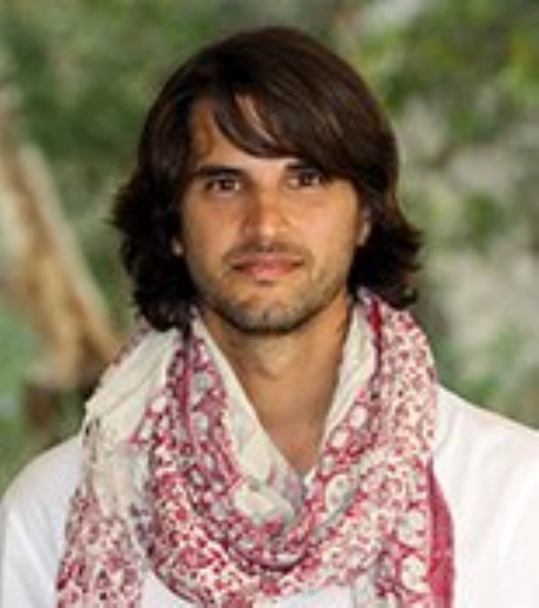}}]{Emre Neftci}
Dr. Emre Neftci received his M.Sc. degree in physics from Ecole Polytechnique Federale de Lausanne, Switzerland, and his Ph.D. in 2010 at the Institute of Neuroinformatics at the University of Zurich and ETH Zurich. Currently, he is an assistant professor in the Department of Cognitive Sciences and Computer Science at the University of California, Irvine. His current research explores the bridges between neuroscience and machine learning, with a focus on the theoretical and computational modeling of learning algorithms that are best suited to neuromorphic hardware and non-von Neumann computing architectures.
\end{IEEEbiography}

\end{document}